# A Semi-Supervised Method for Predicting Cancer Survival Using Incomplete Clinical Data

Hamid Reza Hassanzadeh, *IEEE Member*, John H. Phan, *IEEE Member*, and May D. Wang, *IEEE Senior Member*

*Abstract*— Prediction of survival for cancer patients is an open area of research. However, many of these studies focus on datasets with a large number of patients. We present a novel method that is specifically designed to address the challenge of data scarcity, which is often the case for cancer datasets. Our method is able to use unlabeled data to improve classification by adopting a semi-supervised training approach to learn an ensemble classifier. The results of applying our method to three cancer datasets show the promise of semi-supervised learning for prediction of cancer survival.

## I. Introduction

Prediction of survival for cancer patients is a challenging task; however, it is important for determining treatment course. Physicians typically make subjective decisions based on past experience. These decisions can also prevent unnecessary therapy and improve patient quality of life [1]. Moreover, the prediction of survival can enhance the ability to determine correct prognosis which, in itself, is valuable for research and timely referral of patients to hospice care [2].

Omics data (i.e., genomics and proteomics) have been a promising source of information for identifying molecular signatures of cancer. Thus, different strategies have been developed [3-6] to exploit high-throughput data for the prediction of cancer survival. However, these approaches pose several challenges that should be addressed before a comprehensive and accurate prediction can be achieved. Some of these challenges include: lack of a biological understanding of the related pathways and genes, data scarcity, and existence of a plethora of irrelevant features which altogether, result in difficulty in training accurate omics-based classifiers.

Yet a successful surrogate strategy for predicting survival is based on using clinical factors that are recorded through the course of a patient's treatment. The issue with such data is that it often contains missing values.

To date, there have been several approaches suggested in the literature which differ mostly in the adopted data mining technique and how to deal with the missing attribute values and labels. An important shortcoming that a majority of these methods share is that they are either designed for big datasets or have not been tested on such datasets. Snow et al. [7] used an artificial neural network (ANN) to predict 5-year survival after colon carcinoma treatment. They selected a training set of patients that either survived for more than 5 years after treatment or died within 5 years. In a similar study, Ng et al. [8] developed an ANN to predict the survival time of terminally ill cancer patients using clinical data from National Cancer Center Singapore (NCCS). To deal with the missing data, they either removed the attributes with missing values or excluded patients with missing values. In [9], authors maintained that classifying the samples in a crisp way may not always be the best choice and suggested a fuzzy-based decision tree approach. They applied their method to a breast cancer dataset and removed the cases with missing attribute values. Thongkam et al. [10] proposed another approach to predict breast cancer survivability using AdaBoost. They only kept patients who either survived for more than five years or died within that period. Finally, Abreu et al. [11] used three different ensemble methods to predict survival of breast cancer patients and used k-nearest neighbors to impute missing values. Moreover, they also filtered out patients who were still alive at the time of the study but who had not lived more than the threshold of the study.

A prediction approach suitable for small datasets should address several key challenges, as follows:

- It should be able to deal with incomplete data since filtering out the incomplete samples would prohibitively reduce the size of the dataset.
- It must generalize well to avoid over-fitting.
- It should account for the heterogeneity in the data. For instance, it should remain robust when predicting survival in patients with different histologic grades.
- It should be able to use all available data. A large number of patients in the datasets are those who dropped out of the study before the survival threshold. These patients are typically unlabeled samples that are not included for classification. While we do not know the label for these samples, there may be useful information in these patient samples that may improve the accuracy of the prediction model.

We present a method for predicting survival of cancer patients using incomplete clinical data. The novelty of this method is two-fold. First, this method is able to deal with noisy

This research has been supported by grants from National Institutes of Health (Center for Cancer Nanotechnology Excellence U54CA119338, and R01 CA163256), Georgia Cancer Coalition (Distinguished Cancer Scholar Award to Professor Wang).

H. R. Hassanzadeh is with the Department of Computational Science and Engineering, Georgia Institute of Technology, Atlanta, GA 30332 USA. (e-mail: hassanzadeh@gatech.edu).

J. H. Phan is with the Department of Biomedical Engineering, Georgia Institute of Technology and Emory University, Atlanta, GA 30332 USA (e-mail: jhphan@gatech.edu).

M. D. Wang is with the Department of Biomedical Engineering, Georgia Institute of Technology and Emory University and the School of Electrical and Computer Engineering, Georgia Institute of Technology, Atlanta, GA 30332 USA (corresponding author, phone: 404-385-2954; e-mail: maywang@bme.gatech.edu).

TABLE I. NUMBER OF LABELED VS. UNLABELED SAMPLES

| Cancer Sub-Type | # of Labeled Samples | | # Un-Labeled Samples | % Un-Labeled |
|---|---|---|---|---|
| | #Positive | #Negative | | |
| KIRC (5-y survival) | 111 | 142 | 279 | 52% |
| OV (3.5-y survival) | 180 | 197 | 210 | 36% |
| PAAD (1.4-y survival) | 38 | 36 | 100 | 57% |

TABLE II. STATISTICS OF MISSING ATTRIBUTES FOR ORIGINAL AND PROCESSED CLINICAL DATASETS

| Cancer Sub-Type | Unprocessed Clinical Data | | Processed clinical data | |
|---|---|---|---|---|
| | # Factors | % Missing Values | # Retained Factors | % Missing Values |
| KIRC | 55 | 31.38% | 16 | 2.5% |
| OV | 50 | 33.8% | 10 | 3.3% |
| PAAD | 70 | 38.58% | 26 | 23.49% |

input data and is robust to over-fitting caused by data scarcity. Second, this method uses a semi-supervised learning paradigm that leverages available unlabeled data to improve prediction accuracy.

The remainder of this paper is organized as follows. In section II, we present our method and describe three TCGA datasets used to evaluate our method. In section III, we present our results which show that the semi-supervised approach can exploit the unlabeled samples to improve the accuracy of the prediction. Finally, we conclude in section IV.

## II. MATERIAL AND METHODS

### A. Clinical Cancer Datasets

We use the clinical data associated with cancer patients from the Cancer Genome Atlas (TCGA). Specifically, we are interested in three important sub-types of cancers, namely, the kidney or renal cell carcinoma (KIRC), ovarian serous cystadenocarcinoma (OV) and, the pancreatic ductal adenocarcinoma (PAAD). Table I lists the statistics of these datasets.

### B. Data Pre-Processing

We use the following protocol to pre-process the data to make them suitable for prediction modeling.
- Any attribute that has more than four possible assignments is stratified into at most four categories.
- Attributes that have the same value for more than 99% of the patients are removed.
- For each dataset and a given survival threshold, we divide the samples into two groups, namely, the labeled and the unlabeled groups. The labeled category includes those patients who did not survive for the threshold amount of years (negative class) or who are known to have lived for at least that amount of time (positive class). The unlabeled data includes patients still alive at the time of last follow-up, but not known to have lived more than the specified threshold.
- Survival thresholds are selected such that almost half of the labeled samples fall within the positive class and the rest in the negative class.

Table I lists the size of each category for each cancer dataset and the relative size of unlabeled samples. Table II lists attribute statistics of the datasets before and after pre-processing. According to the table, kidney and ovarian cancers show relatively similar statistics except that, compared to the kidney cancer dataset, a smaller number of informative features are retained for ovarian cancer. In contrast, the pancreatic cancer dataset exhibits a significant difference in terms of the percentage of missing entries and the size of the available dataset. Table III lists the set of attributes and their corresponding possible value assignments for the kidney, ovarian, and pancreatic cancer datasets, respectively.

TABLE III. LIST OF PROCESSED CLINICAL FEATURES

| Renal Cancer | Ovarian cancer | Pancreatic cancer |
|---|---|---|
| 1. Race | 1. Race | 1. Race |
| 2. Gender | 2. Ethnicity | 2. Gender |
| 3. Neoplasm Histologic Grade | 3. Initial Pathologic Diagnosis Method | 3. Prior Dx |
| 4. Laterality | 4. Person Neoplasm Cancer Status | 4. Initial Pathologic Diagnosis Method |
| 5. Tissue Prospective Collection Indicator | 5. Neoplasm Histologic Grade | 5. Surgery Performed Type |
| 6. Tissue Retrospective Collection Indicator | 6. Venous Invasion | 6. Neoplasm Histologic Grade |
| 7. Pathologic T | 7. Lymphatic Invasion | 7. Histologic Grading Tier Category |
| 8. Pathologic N | 8. Age | 8. Maximum Tumor Dimension |
| 9. Pathologic M | 9. Anatomic Neoplasm Subdivision | 9. Age |
| 10. Primary Lymphnode Presentation Assessment | 10. Pathologic Stage | 10. Residual Tumor |
| 11. Person Neoplasm Cancer Status | | 11. Pathologic T |
| 12. Prior Dx | | 12. Pathologic N |
| 13. Pathologic Stage | | 13. Pathologic M |
| 14. History of Neoadjuvant Treatment | | 14. Pathologic Stage |
| 15. Age | | 15. Person Neoplasm Cancer Status |
| 16. Hemoglobin Result | | 16. Tobacco Smoking History |
| | | 17. History of Diabetes |
| | | 18. History of Chronic Pancreatitis |
| | | 19. Family History of Cancer |
| | | 20. Radiation Therapy |
| | | 21. New Tumor Event after Initial Treatment |
| | | 22. Anatomic Neoplasm Subdivision |
| | | 23. Targeted Molecular Therapy |
| | | 24. Alcohol Consumption Frequency |
| | | 25. Tissue Prospective Collection Indicator |
| | | 26. Tissue Retrospective Collection Indicator |

### C. Ensemble Classification

We use ensemble learning because it is robust to over-fitting and combines multiple classifiers, each of which performs well for a specific part of the input space. The net outcome of ensemble learning results from assigning higher weights to weak learners that perform better for the given input samples. Specifically, we use Robust Boost [12], which has been shown to be robust to noisy labels. This robustness is desirable when adding samples to the training set as a result of semi-supervised training. Moreover, we use decision trees from the classification and regression tree (CART) toolbox that is able to deal with missing attribute values by using a technique called surrogate splitting [13].

### D. Semi-Supervised Prediction Modeling

We propose a semi-supervised approach that leverages unlabeled training data to increase sample size (Figure 1). After pre-processing, we use the initial labeled data to train a classifier and predict the labels of the unlabeled data. Next, we choose a confidence threshold according to which we can select part of the unlabeled set that can potentially improve the labeled sample size. We only use unlabeled samples that the model classifies with a high confidence level. In other words, the prediction score of these unlabeled samples exceed some threshold. We find this threshold by means of a 10-fold cross-validation on the training set. More specifically, for each fold we record the threshold that results in the highest prediction accuracy after running the whole pipeline. To choose the final threshold, we use a majority voting among these recorded thresholds that were found for each fold. Once we find the threshold, we apply the learned model iteratively to the remainder of the samples in the unlabeled set. During each round of prediction we choose the samples whose prediction score is beyond the confidence threshold and move them to the training set for the next round. We label these samples according to the classifier's prediction. This iterative procedure continues until no prediction passes the confidence test in which case we train our model on the compiled training set to generate the final model.

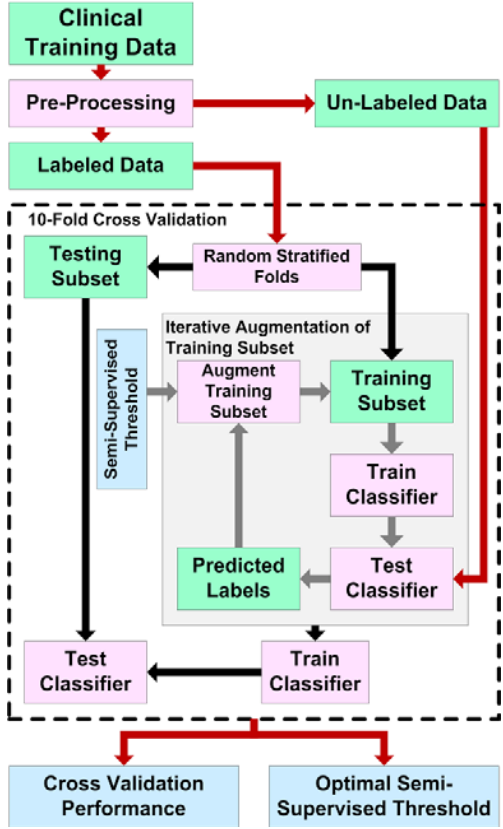

Figure 1. Optimization of the optimal semi-supervised threshold using cross validation.

## III. RESULTS AND DISCUSSION

We evaluate the performance of the prediction model when trained using supervised learning and semi-supervised learning paradigms. We use accuracy (Acc), mean of sensitivity and specificity (SnSp/2), and the Matthews correlation coefficient (MCC) [14] as our performance measures. All evaluation metrics are computed using 5-fold cross validation.

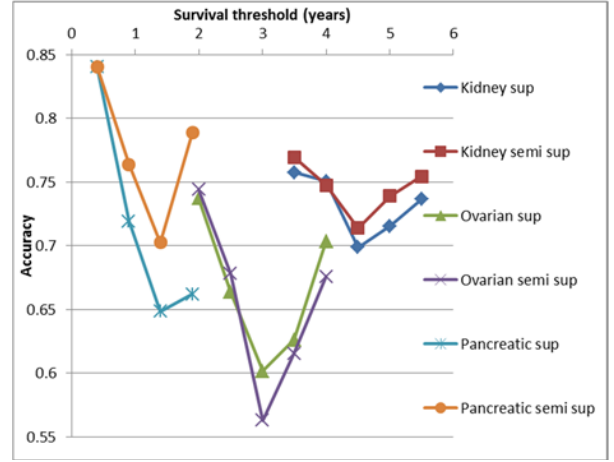

Figure 2. Comparison of prediction performance for supervised (sup) and semi-supervised (semi sup) training for kidney, ovarian, and pancreatic cancers. Models were compared for different survival thresholds.

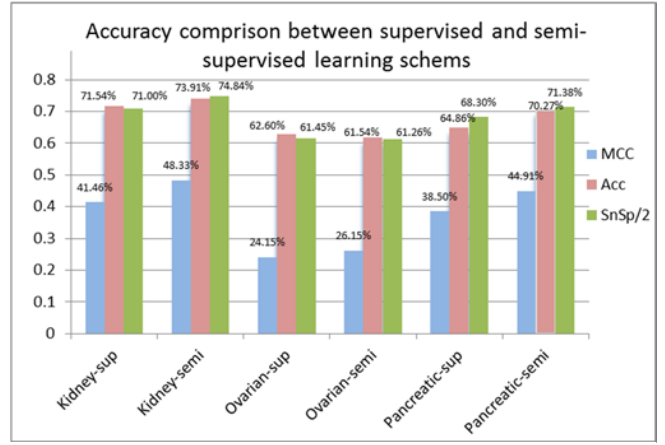

Figure 3. Performance comparison between supervised and semi-supervised methods at fixed survival thresholds of 5, 3.5, and 1.4 years for kidney, ovarian, and pancreatic cancer, respectively.

Figure 2 shows the prediction accuracy of the trained model for both supervised and semi-supervised learning paradigms. As clearly seen from the figure, the semi-supervised approach is superior to the supervised approach for the kidney and pancreatic cancer datasets. This performance is more pronounced in the latter case due to the smaller size of the dataset as evidenced by Table I. This corroborates our previous assertion that semi-supervised training can offer considerable benefits when data are scarce. For the ovarian dataset, supervised learning performs slightly better for some survival thresholds. This is expected due to the smaller number of available informative predictors (Table II and Table IV). Furthermore, Figure 3 compares the 5, 3.5 and, 1.4-year survival prediction performance measures for the kidney, ovarian, and pancreatic cancers, respectively. According to this figure, even though the accuracy of the supervised approach applied to the ovarian cancer is slightly higher than the semi-supervised counterpart, the corresponding MCC is still smaller, which suggests that higher accuracy of the supervised approach may be related to the distribution of the positive and negative classes rather than to the learning method itself.

TABLE IV. IMPORTANT FEATURES OF FINAL CLINICAL PREDICTORS

| Kidney cancer | | Ovarian cancer | | Pancreatic cancer | |
|---|---|---|---|---|---|
| Factor | Importance | Factor | Importance | Factor | Importance |
| neoplasm cancer status | 0.77 | neoplasm cancer status | 0.805 | molecular therapy | 0.2419 |
| lymphatic invasion | 0.0904 | age | 0.1634 | age | 0.1418 |
| age | 0.0575 | clinical stage | 0.018 | pathologic T | 0.1222 |
| clinical stage | 0.0411 | lymphatic invasion | 0.0137 | pathologic stage | 0.1088 |
| histologic grade | 0.0281 | | | radiation therapy | 0.0858 |
| race | 0.013 | | | residual tumor | 0.083 |
| | | | | histologic grade | 0.076 |
| | | | | tumor dimension | 0.0583 |
| | | | | anatomic neoplasm subdivision | 0.0416 |
| | | | | surgery type | 0.0407 |

Finally, Table IV lists the most important features and their corresponding weights found by our trained model. Specifically, these are the importance degrees of the decision trees averaged over all weak learners. There are a few interesting facts implied by this table. First, tumor-related attributes, if included in the feature set, account for most of the predictive power of the model. Furthermore, age of the patient is another important factor that contributes to the prediction of patient survival. Last, race is a factor in kidney cancer survival. This is likely related to genetic factors that are different among different populations.

## IV. CONCLUSION AND FUTURE WORK

We designed a classifier that was able to predict labels for unlabeled data with missing attribute values. We then used a semi-supervised learning approach to train a prediction model and showed that, for the task of survival prediction in the presence of a significant amount of unlabeled data, semi-supervised learning can improve performance. We applied the semi-supervised learning approach to kidney, ovarian, and pancreatic cancer data. Moreover, the weights of the features in the trained model can be interpreted by human experts.

We simplified the survival prediction problem by converting the original regression problem into a binary classification problem using a survival threshold. However, this removes information that may be used to further improve accuracy. Future approaches may consider using semi-supervised learning in the context of regression to predict cancer survival.

## V. ACKNOWLEDGEMENT

The authors thank Janani Venugopalan and Dr. James Cheng for assisting in manuscript preparation.

## REFERENCES

[1] P. Glare, K. Virik, M. Jones, M. Hudson, S. Eychmuller, J. Simes, *et al.*, "A systematic review of physicians' survival predictions in terminally ill cancer patients," *British Medical Journal,* vol. 327, pp. 195-198, Jul 26 2003.

[2] S. Gripp, S. Moeller, E. Bolke, G. Schmitt, C. Matuschek, S. Asgari, *et al.*, "Survival prediction in terminally ill cancer patients by clinical estimates, laboratory tests, and self-rated anxiety and depression," *Journal of Clinical Oncology,* vol. 25, pp. 3313-3320, Aug 1 2007.

[3] S. Michiels, S. Koscielny, and C. Hill, "Prediction of cancer outcome with microarrays: a multiple random validation strategy," *Lancet,* vol. 365, pp. 488-92, Feb 5-11 2005.

[4] A. Prat, A. Lluch, J. Albanell, W. T. Barry, C. Fan, J. I. Chacon, *et al.*, "Predicting response and survival in chemotherapy-treated triple-negative breast cancer," *British Journal of Cancer,* vol. 111, pp. 1532-1541, Oct 14 2014.

[5] X. Chen and L. Wang, "Integrating biological knowledge with gene expression profiles for survival prediction of cancer," *J Comput Biol,* vol. 16, pp. 265-78, Feb 2009.

[6] D. Kim, J. G. Joung, K. A. Sohn, H. Shin, Y. R. Park, M. D. Ritchie, *et al.*, "Knowledge boosting: a graph-based integration approach with multi-omics data and genomic knowledge for cancer clinical outcome prediction," *J Am Med Inform Assoc,* vol. 22, pp. 109-20, Jan 2015.

[7] P. B. Snow, D. J. Kerr, J. M. Brandt, and D. M. Rodvold, "Neural network and regression predictions of 5-year survival after colon carcinoma treatment," *Cancer,* vol. 91, pp. 1673-8, Apr 15 2001.

[8] T. Ng, L. Chew, and C. W. Yap, "A Clinical Decision Support Tool To Predict Survival in Cancer Patients beyond 120 Days after Palliative Chemotherapy," *Journal of Palliative Medicine,* vol. 15, pp. 863-869, Aug 2012.

[9] M. U. Khan, J. P. Choi, H. Shin, and M. Kim, "Predicting breast cancer survivability using fuzzy decision trees for personalized healthcare," in *Engineering in Medicine and Biology Society, 2008. EMBS 2008. 30th Annual International Conference of the IEEE*, 2008, pp. 5148-5151.

[10] J. Thongkam, G. D. Xu, and Y. C. Zhang, "AdaBoost Algorithm with Random Forests for Predicting Breast Cancer Survivability," *2008 Ieee International Joint Conference on Neural Networks, Vols 1-8,* pp. 3062-3069, 2008.

[11] P. Abreu, H. Amaro, D. Silva, P. Machado, M. Abreu, N. Afonso, *et al.*, "Overall Survival Prediction for Women Breast Cancer Using Ensemble Methods and Incomplete Clinical Data," in *XIII Mediterranean Conference on Medical and Biological Engineering and Computing 2013*. vol. 41, L. M. Roa Romero, Ed., ed: Springer International Publishing, 2014, pp. 1366-1369.

[12] Y. Freund, "A more robust boosting algorithm," *arXiv preprint arXiv:0905.2138,* 2009.

[13] L. Breiman, J. Friedman, C. J. Stone, and R. A. Olshen, *Classification and regression trees*: CRC press, 1984.

[14] B. W. Matthews, "Comparison of the predicted and observed secondary structure of T4 phage lysozyme," *Biochimica et Biophysica Acta (BBA)-Protein Structure,* vol. 405, pp. 442-451, 1975.